# A Class-Incremental Learning Method Based on One Class Support Vector Machine

Chengfei Yao*, Jie Zou, Yanan Luo, Tao Li, Gang Bai**
College of Computer and Control Engineering
Nankai University
Tianjin, China
E-mail: *2120160400@mail.nankai.edu.cn, **baigang@nankai.edu.cn

*Abstract*—A method based on one class support vector machine (OCSVM) is proposed for class incremental learning. Several OCSVM models divide the input space into several parts. Then, the 1VS1 classifiers are constructed for the confuse part by using the support vectors. During the class incremental learning process, the OCSVM of the new class is trained at first. Then the support vectors of the old classes and the support vectors of the new class are reused to train 1VS1 classifiers for the confuse part. In order to bring more information to the certain support vectors, the support vectors are at the boundary of the distribution of samples as much as possible when the OCSVM is built. Compared with the traditional methods, the proposed method retains the original model and thus reduces memory consumption and training time cost. Various experiments on different datasets also verify the efficiency of the proposed method.

## I. INTRODUCTION

Incremental learning is a hot research topic of pattern classification. Incremental learning fully retains the learned classification knowledge and modifies the original model using the knowledge learned from the new samples. The great significance of incremental learning is that when new samples are added, there is no need to retrain with all samples, reducing memory consumption and training time cost. In terms of the classes which the new samples belong to, incremental learning can be divided into two types. The common type of incremental learning is that the new training samples belong to the old classes. Another is the new training samples belong to the new class, called class incremental learning (CIL). Researchers have achieved rich results in the former [1] [2]. Research on CIL is limited. CIL needs to establish a learning strategy to deal with the new classes joined, while retaining the original model as much as possible. CIL has a wide range of applications in the classification scene that the number of classes changes dynamically, such as text classification, sound classification. In this paper, we focus on the multi-classification strategy of CIL.

Some researchers have made attempt to achieve CIL. [3] proposed a class incremental learning method for SVM. One binary classifier is constructed by using the new samples and all the old samples, used to determine whether test samples belong to the new class or the old classes. This method reduces training time successfully when new classes are added. However, as the number of classes increases, the number of samples in the newly added class and the old classes becomes very unbalanced, resulting in poor performance. Moreover, this method can not build model when there is only one class sample. Therefore, some researchers are devoted to solving class incremental learning by establishing multi-class classifiers through several one-class classifiers [4] [5] [6]. [4] proposed a class incremental learning method called hyper-sphere class incremental learning (HS-CIL). HS-CIL trains several hyper-spheres and decides its label by measuring the distance between test samples and each hyper-sphere center. However, this method lacks consideration of the true distribution density of samples and non-convex distribution of samples. [5] proposed a method for multi-classification based on OCSVM. This method trains several OCSVM models and uses the defined dissimilarity to determine the ownership of the test samples. However, each OCSVM model is in different mapped space because of the different Gaussian kernel width and the calculated values are incomparable. [6] proposed a mahalanobis hyperellipsoidal learning machine class incremental learning algorithm (MHE-CIL), but this method also did not perform well when the distribution of samples is not convex in the feature space.

In this paper, we propose a class incremental learning method based on OCSVM. The basic idea is that some OCSVM models divide the input space into several parts, which can be controversial and non-controversial parts. If only one OCSVM predicts the test sample is positive, the part located by this test sample is non-controversial. Otherwise it is controversial. The hyper-plane of each OCSVM can be imagined as an enclosed surface of the target class samples in the input space. If the enclosed surfaces are suitable, the surfaces fit well with the shape of the samples distribution, and the support vectors lie on the boundary of samples distribution.The support vectors are used to build 1VS1 classifiers for handling disputes. When a new class comes, what needs to be done is to train the OCSVM for this class and solve the disputes between this class and old classes. The main contribution of our method is to make the classification process easier to explain by introducing OCSVM into class incremental learning. Another contribution is to make the support vectors of OCSVM carry more information by adjusting parameters.

The reminder of this paper is organized as follows. Section II briefly introduces OCSVM and a parameter selection method for OCSVM. Section III makes a complete presentation of the proposed method. Section IV compares our method with the traditional classifiers on both toy datasets and

UCI datasets. Finally, section V summarizes the paper and looks forward to future work.

## II. RELATED WORK

Because new samples of class incremental learning belong to a new class, we use OCSVM to learn the distribution of new class. The principle of OCSVM is as follows.

### A. OCSVM

One-class classification builds model using only positive samples, and determine whether test samples belong to the training class or not. One-class SVM is proposed by B Schölkopf and J Platt in 2001 to solve OCC [7]. One-class support vector machine find a hyper-plane in high-dimensional space to separate the target sample and the origin as far as possible. The hyper-plane can be imagined as an enclosed surface of the target sample in the input space, just similar to the minimum enclosing ball [8]. As other classifiers of SVM family, OCSVM has strong generalization ability because of its minimized structural risk, but sparse data storage and a reasonable training cost.

To separate the data set from the origin, we should solve the following program:

$$\min_{\omega \in F, \xi \in \mathbb{R}^l, \rho \in \mathbb{R}} \frac{1}{2}\|\omega\|^2 + \frac{1}{vl}\sum_i \xi_i - \rho \quad (1)$$

subject to $(\omega \cdot \phi(\mathbf{x}_i)) \geq \rho - \xi_i, \xi_i \geq 0$

Here, the parameter ν is an upper bound on the fraction of training samples lying on the wrong side of the hyper-plane and the lower bound of the fraction of all support vectors.

The dual problem (2) of OCSVM problem is obtained by using the Lagrange multiplier method.

$$\max_{\alpha} -\frac{1}{2}\sum_{i,j=1}^{n} \alpha_i \alpha_j k(\mathbf{x}_i, \mathbf{x}_j) \quad (2)$$

subject to $0 \leq \alpha_i \leq \frac{1}{vl}, \sum_{i=1}^{n} \alpha_i = 1$

The decision function of OCSVM is as following:

$$f(\mathbf{x}) = \text{sgn}\left(\sum_{i=1}^{n} \alpha_i k(\mathbf{x}_i, \mathbf{x}) - \rho\right) \quad (3)$$

If the sample's decision value is greater than 0, it is judged that the sample belongs to the class, otherwise it does not belong to the class.

### B. MIES

Parameter selection of one-class SVM, especially kernel parameter, is significant just like many other classifiers. There are many attempts to select fine-tuned parameters for one-class SVM with RBF kernel function [9] [10] [11]. MIES algorithm is a parameter selection algorithm of Gaussian kernel for one-class SVM based on geometric essence [12]. The basic idea is based on the observation that the hyper-plane of feature spaces is imagined as the enclosing surface of the input space. If the surfaces are appropriate, they should be as close as possible to the boundary point and as far as possible from the inner point.

MIES algorithm achieves the above idea by measuring the distance between the enclosing surface and the internal point and the boundary point. Equation (4) is the objective optimization function of the MIES algorithm. Among the candidate parameters, the parameter that maximizes the equation is selected. In order to compare the distances in different spaces determined by different parameter s, the distances have been normalized.

$$f_0(s) = \max_{\mathbf{x}_i \in \Omega_{IN}} d_N(\mathbf{x}_i) - \max_{\mathbf{x}_j \in \Omega_{ED}} d_N(\mathbf{x}_j) \quad (4)$$

The MIES algorithm relies on boundary points and internal points, so it is necessary to select internal points and boundary points from only one class sample automatically. The author adds the border-edge pattern selection method (BEPS) proposed by Yuhua Li and Liam Maguire for selecting boundary points based on local geometric information into MIES [13]. The basic idea of BEPS algorithm is that if a point is a boundary point, then most of its neighbors should be on the side of its tangent plane.

Because of the optimization goal of the MIES algorithm, the OCSVM with parameters determined by the MIES algorithm has a good geometric interpretation in the input space.

## III. OUR METHOD

The basic idea of our method is that several OCSVM models divide the input space into two types of subspace, the controversial part and the non-controversial one. First, the OCSVM model of each class is built. If only one OCSVM predicts that the test sample is positive, the part is non-controversial. Otherwise it is controversial.

### A. Train OCSVM

In the proposed method, the role of OCSVM is crucial. It mainly estimates the distribution of samples of each class and determines whether the test sample belongs to this class. To establish an OCSVM with Gaussian kernel, it is mainly necessary to determine two parameters s and ν. The parameter s is the width of Gaussian kernel for OCSVM model, and the parameter ν mainly controls the fraction of samples against hyperplane in high dimensional feature space, as shown in Figure 1. The value of the parameter ν is generally taken as the empirical threshold among the interval of [0.1-0.4]. Because the MIES algorithm is based on geometric essence, the support vectors can be roughly at the boundary. However, due to the contradiction between the two optimization objectives, the distribution of support vectors can sometimes be uneven. We make some changes on MIES to make it more responsive to the proposed method.

Since the strategy of resolving disputes is based on the support vectors, the distribution of support vectors directly affects the performance of the method. In general, when the support vectors are on the boundary of each class, the shell formed by the support vectors is the best represent for this class. In paper [13], the author also use boundary and border samples to achieve the same classification performance as all samples.

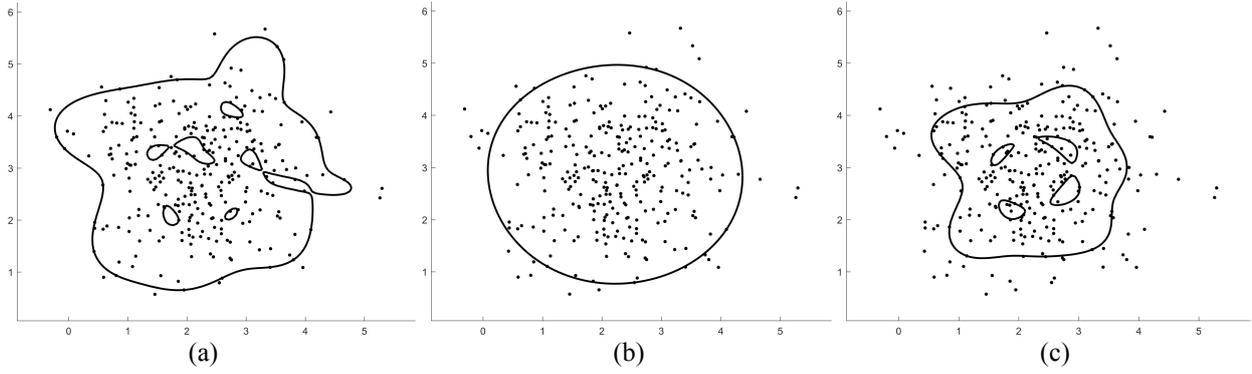

Fig. 1. OCSVM model with different parameter v and s. (a) v=0.1, s=1. (b) v=0.1, s=10. (c) v=0.3, s=1.

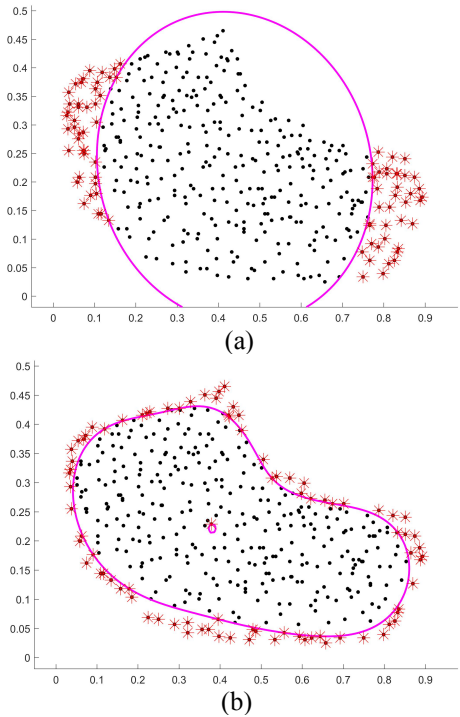

Fig. 2. the comparison of support vectors for MIES and modified MIES. (a) MIES. (b) Modified MIES.

The optimization goal of the MIES algorithm is to make the enclosing surface as close as possible to the boundary samples and as far as possible from the internal samples. The support vectors obtained are roughly distributed on the boundary. However, sometimes the above two goals can not be satisfied at the same time. In fact, MIES does not require that the enclosing surface be far from every internal point. Therefore, in order to get the required support vectors, we change the optimization goal of parameter selection of Gaussian kernel to be as close as possible to the boundary point. The objective optimization function is as following.

$$s = \arg\min\left(\max_{\mathbf{x}_i \in \Omega_{ED}} d_N(\mathbf{x}_i)\right) \quad (5)$$

In fact, as long as almost all of the selected boundary samples are support vectors, the optimization function can be satisfied. We limit the ratio of support vectors to prevent all samples from becoming support vectors as the value of parameter s is small. With a large number of experiments, this ratio is preferably within the range of v to 1.5v. As shown in Fig2, the support vectors obtained by the modified MIES are more evenly distributed than the support vectors obtained by the MIES on the sample distribution boundary. The complete process of training OCSVM is as follows.

| Algorithm 1: Train OCSVM |
|---|
| **Input**: the training set, v |
| Determine the parameter of Gaussian kernel according to modified MIES: |
|     Select the boundary point from the training set according to BEPS algorithm |
|     Exclude the candidates of s if the fraction of support vector exceeds [v, 1.5 *v] |
|     Select s according to (6) from the left candidates of s |
| Train OCSVM based on parameter v and s |
| **Output**: the OCSVM model |

### B. Reuse SV to Train Multi-Classification Model

If there is no overlapping between the categories, OCSVM can accurately determine the label of test sample. In practice, the distribution of different classes often have overlapping parts, so we need to establish a strategy to deal with the disputes. The support vectors obtained according to Algorithm 1 are mostly boundary samples. These support vectors contain the main information needed for multi-classification. We reuse these support vectors to establish a strategy for the controversial part. The strategy to solve the disputes is to establish a 1VS1 classifier. In this paper, the original binary SVM is selected.

Then when a new class comes, what needs to do is to train the OCSVM for this class and then train the 1VS1 classifiers by using this class's support vectors and the support vectors of the old classes.

In the testing phase, first the OCSVM is used to determine whether the test sample falls into the controversial part or not. In all OCSVM predictive labels, if only one is positive, the test sample is considered to be in the uncontroversial part. If more than one or none is positive, then it is considered to be in the disputed part. When the test sample is in the non-controversial part, the unique class label is directly assigned to the test sample. When the test sample is in the controversial part, the first step is to get the classes to which the test sample may belongs. Then support vectors of these classes are used to predict the label of test sample according to the corresponding 1VS1 classifiers. Because the method is based on the distribution of samples, we call it SD-CIL. The complete flow of SD-CIL is as follows.

**Algorithm 2: the SD-CIL method**
**Training phase:**
1. Train OCSVM for each class according to Algorithm 1.
2. Train binary SVM for the different combinations of classes.

**Testing phase:**
1. Send the test sample into all OCSVM for testing.
2. If only one prediction is positive, the test sample belongs to this class.
   Elseif more than one is positive, regard a class that has a positive label as a possible class.
      Get the label of the test sample using binary SVM of the possible classes. (The parameters are determined by cross-validation.)
   Else none is positive, regard all classes as possible classes.

## IV. EXPERIMENT

In order to test the performance of the proposed method, experiments on toy datasets and UCI datasets are conducted. The basic information of these datasets is shown in Table 3. We compare the performance of the proposed method with SVM, KNN and the multi-classification method based on OCSVM in [3] (abbreviated as OC2MC). The method that NN is selected as the strategy for the disputes is also added as a comparison (abbreviated as OCSVM-NN).

### A. Experements on Toy Datasets

Because the classification results can be easily visualized on two-dimensional data sets, experiments are first conducted on two-dimensional datasets. We generate several two-dimensional datasets of different shapes to test the performance of the proposed method in multi-classification. The classification map identified by the proposed method and other methods are compared. The results are shown in Figs 3-4. In the subfigure (d) (e), the support vectors of different classes are marked. Obviously, the modified MIES algorithm can make the support vectors distribute on the boundary of the distribution of samples. These support vectors contain the information required for multi classification.

TABLE I. THE BASIC INFORMATION OF DATASETS

| Dataset | Dimension | Training patterns | Test patterns |
|---|---|---|---|
| Toy Dataset1 | 2 | 783 | 336 |
| Toy Dataset2 | 2 | 954 | 408 |
| Iris | 4 | 105 | 35 |
| Seeds | 7 | 147 | 63 |
| Pima | 8 | 538 | 230 |
| Waveform | 21 | 3500 | 1500 |
| Transfusion | 4 | 524 | 224 |

Subfigure (e) represents for the classification map determined by our method, (b) (c) (d) (f) for other methods. The classification map of the proposed method is similar to the classification map determined by SVM, which is better than other methods. The classification map of the proposed method is made up of two parts. The uncontroversial part is determined by the OCSVM model and the controversial part is derived from the binary classifier. The results show that this composition is reasonable.

Table 2 shows the mean accuracy and standard deviation of different methods on these two datasets. For each dataset, we perform the different methods 10 times. In each experiment, the dataset is randomly divided. 70% of each class used for training, and the remaining 30% is used for testing. The result shows that our method has achieved competitive accuracy.

### B. Experements on UCI Datasets

To test the performance of the proposed method in practice, we chose different datasets from the UCI datasets. The basic information of these datasets is shown in Table 1. We run 10 times on each dataset to compare the accuracy of different methods. Also, the datasets are randomly divided each time. 70% of each classes used for training, and the remaining 30% is used for testing.

The parameters of our method are set as follows. On the Pima and Seeds dataset, the value of parameter v is 0.3 and the value is 0.2 on remaining datasets. The parameter s is determined by the modified MIES algorithm automatically. The parameters of the remaining methods have been fine-tuned. The results are shown in Table 2. The results show that our method works well across all datasets. And a large number of test samples in each dataset get the correct label by OCSVM. The results show that it is reasonable to divide the input space into two parts: the controversial part and the uncontroversial part. On the other hand, as the number of dimensions increases, the accuracy of the OCSVM-1NN decreases. The main reason is that the NN strategy has higher requirements for the distribution of support vectors. In high dimensions, it is difficult to make the support vectors at the boundary than in the low dimension.

TABLE II. CLASSIFICATION RESULTS ON DIFFERENT DATASETS

|              | SD-CIL          | SVM             | 1NN             | OCSVM-1NN       | OC2MC           |
|--------------|-----------------|-----------------|-----------------|-----------------|-----------------|
| Toy Dataset1 | 99.49 ± 0.003   | 99.38 ± 0.220   | 98.78 ± 0.004   | 98.54 ± 0.003   | 99.34 ± 0.003   |
| Toy Dataset2 | 99.26 ± 0.005   | 99.56 ± 0.429   | 99.23 ± 0.004   | 98.50 ± 0.004   | 99.37 ± 0.004   |
| Iris         | 95.78 ± 0.029   | 96.89 ± 2.147   | 95.78 ± 0.024   | 94.22 ± 0.038   | 94.45 ± 0.030   |
| Seeds        | 92.06 ± 0.021   | 90.00 ± 3.819   | 90.16 ± 0.038   | 91.59 ± 0.022   | 91.90 ± 0.026   |
| Pima         | 74.43 ± 0.027   | 77.43 ± 1.861   | 73.91 ± 0.026   | 66.65 ± 0.035   | 74.09 ± 0.031   |
| Waveform     | 85.26 ± 0.006   | 86.75 ± 0.532   | 82.65 ± 0.008   | 74.77 ± 0.012   | 84.63 ± 0.007   |
| Transfusion  | 76.34 ± 0.023   | 77.72 ± 0.879   | 77.86 ± 0.018   | 69.42 ± 0.032   | 65.71 ± 0.031   |

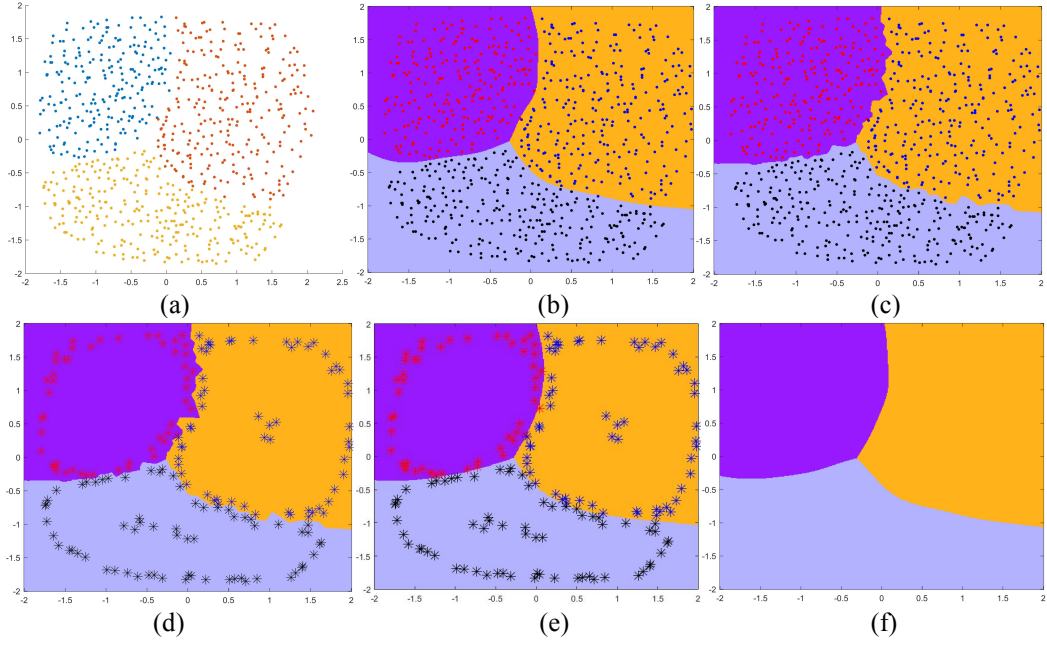

Fig. 3. Performance on Toy Dataset1. (a)Training set. (b) SVM. (c) KNN. (d) OCSVM+KNN. (e) Our method. (f) OC2MC. The "·" of different colors denotes the samples of different classes. The "*" of different colors in (d) (e) represents different support vectors.

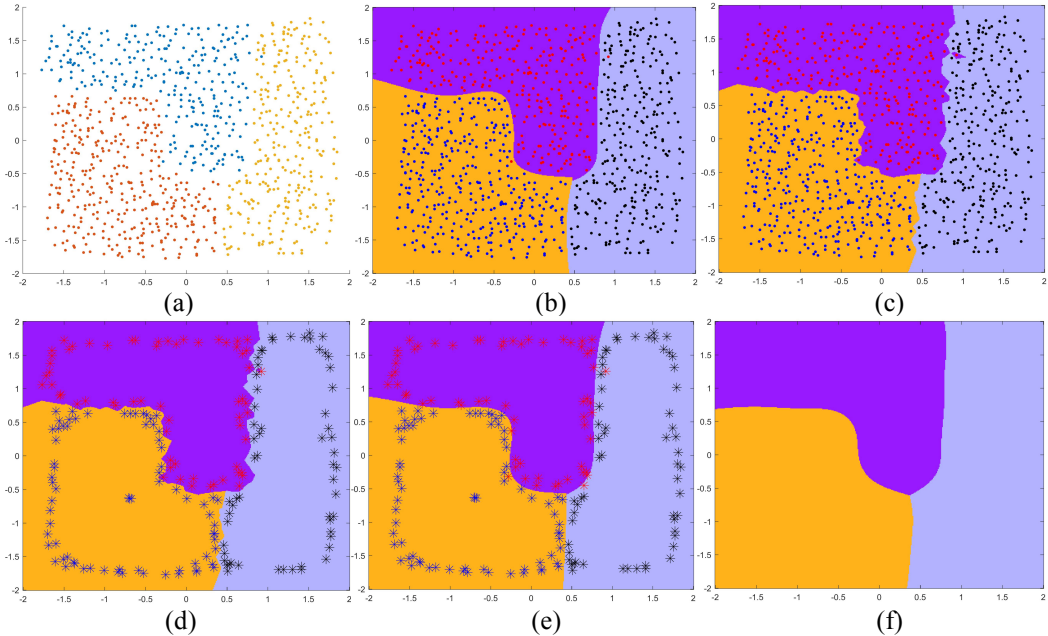

Fig. 4. Performance on Toy Dataset2. (a)Training set. (b)SVM. (c)KNN. (d)OCSVM+KNN. (e)Our method. (f) OC2MC. The "·" of different colors denotes the samples of different classes. The "*" of different colors in (d) (e) represents different support vectors.

## V. DISSCUSION

In this paper, a learning method SD-CIL is proposed for class incremental learning. During the training phase, the distribution of samples is studied by using the OCSVM with RBF kernel. The 1VS1 classifiers to deal with the disputes are built by using support vectors on the boundary. These steps constitute a division of the input space. In the class incremental learning process, there is no need to change the original model, which greatly reduces the training time and memory consumption. During the testing phase, using OCSVM to get the class that the test samples might belong to can also reduce the time of testing, especially for classifiers like KNN. The results show that the proposed method is effective.

Since the 1VS1 classifiers are based on the support vectors on the boundary, it is difficult to solve the classification problem with large overlap. Fortunately, the samples of the overlap in the natural distribution are sparse, and most of them are selected as support vectors. Meantime, how to determine the support vectors in high dimension and make the support vectors form the shell of the distribution of samples in the input space is our future research direction.

ACKNOWLEDGMENT

This work is partially supported by the Natural Science Foundation of Tianjin (No. 16JCYBJC15200, No. 17JCQNJC00300), Tianjin Science and Technology Project (No. 15ZXDSGX00020), the National Key Research and Development Program of China (2016YFC0400709).